\documentclass[letterpaper, 10 pt, conference]{ieeeconf}
\IEEEoverridecommandlockouts                              % This command is only
\makeatletter

\usepackage[utf8]{inputenc} % allow utf-8 input
\usepackage[T1]{fontenc}    % use 8-bit T1 fonts
\usepackage{url}            % simple URL typesetting
\usepackage{booktabs}       % professional-quality tables
\usepackage{amsfonts}       % blackboard math symbols
\usepackage{nicefrac}       % compact symbols for 1/2, etc.
\usepackage{microtype}      % microtypography
\usepackage{xcolor}         % colors
\usepackage{graphicx,epstopdf}

\usepackage{times}
\usepackage{multicol}
\usepackage{hyperref}
\usepackage{bm}
\usepackage{amssymb,amsthm}
\usepackage{amsmath}
\usepackage{mathrsfs}
\usepackage{tabularx}

\usepackage{acronym}
\usepackage{todonotes}
\usepackage{threeparttable}
\usepackage{color}
\usepackage{mathtools}
\usepackage{multirow}
\usepackage[caption=false,font=footnotesize]{subfig}
\usepackage{xcolor} 
\usepackage{soul}
\usepackage{authblk}
\usepackage[utf8]{inputenc} % allow utf-8 input
\usepackage{algorithm}
\usepackage{algpseudocode}
\usepackage{caption}
\title{A Mixed-Reality Testbed for Autonomous Vehicles}
\author{
H. M. Sabbir Ahmad$^{1}$,
Ehsan Sabouni$^{1}$,
Emrullah Celik$^{1}$,
Zean Wan$^{2}$,
Damola Ajeyemi$^{1}$, \\
Christos G. Cassandras$^{1,2}$,
and Wenchao Li$^{1,2}$ \\
$^{1}$ Division of Systems Engineering, Boston University \\
$^{2}$ Department of Electrical and Computer Engineering, Boston University
}

\begin{document}

\maketitle

\begin{abstract}
\label{sec:abstract}
%\textcolor{red}{cbf based controller, use runtime data to learn the parameters online. explain and reposition table II, why idm not included in table II, revise abstract and contribution in introduction, }
%Autonomous driving offers significant benefits in terms of road safety, traffic efficiency, limiting harmful emissions and driver comfort. Despite advancements, existing research predominantly supports Level 2 or Level 3 autonomy due to stringent safety requirements and validation constraints in real-world conditions making this an active area of research and development. Towards this end, 
We propose a mixed-reality, hardware-in-the-loop (HIL) testbed for autonomous vehicles that seamlessly integrates a physical testbed of mobile robots with a high-fidelity simulation environment. The virtual simulation enables the creation of diverse, safety-critical driving scenarios to validate state-of-the-art perception, planning, and control algorithms, while augmenting simulations with physical robots equipped with multimodal sensors in photorealistic virtual environments further facilitating rigorous validation. 
Our testbed also features vehicular connectivity using wireless communication and can accommodate a large number of agents through the combination of physical robots and virtual simulated agents, supporting research on multi-agent systems including Connected and Autonomous Vehicles (CAVs).
Finally, we present a safety-guaranteed framework combining perception, planning and a novel online learning-based controller using Control Barrier Functions (CBFs) for CAVs. Experiments using the proposed framework
are used to validate and demonstrate the key functionalities and the overall utility of the testbed to bridge the gap between simulation and real-world hardware deployment.
\end{abstract}

\section{Introduction}
\label{Introduction}
%Autonomous driving technology including CAVs holds the promise of revolutionizing transportation by enhancing road safety, improving traffic efficiency, and increasing driver comfort. Recent advancements in perception, planning, and control have significantly contributed to the development of autonomous vehicles (AVs). In the realm of perception, deep learning techniques have been employed to interpret complex sensor data, enabling vehicles to understand their surroundings with greater accuracy 
%{\color{blue} Surely there are many PUBLISHED papers on this, rather than using an arXiv document!} 
%\cite{huch2023deepstep, Jaeger2023ICCV, Prakash2021CVPR}. For planning, novel algorithms have been developed to generate optimal trajectories that consider dynamic environments and ensure passenger safety \cite{ye2024path, 9575880}. Control systems have also seen improvements, with the integration of robust control strategies that maintain vehicle stability and performance under various conditions \cite{li2023recent}.

Autonomous Vehicles (AVs) enhance transportation by reducing human error, improving safety, and increasing passenger comfort, while Connected Autonomous Vehicles (CAVs) further amplify these benefits through real-time V2X communication that enables cooperative maneuvers such as platooning and coordinated intersection crossing, thereby improving traffic throughput, reducing congestion, and lowering emissions. Despite these promises, there remains a gap between algorithmic development in simulation and real-world deployment for AVs. Existing simulators such as CARLA~\cite{dosovitskiy2017carla} offer high-fidelity urban environments for algorithm validation. And, to mitigate the ``sim-to-real'' gap, physical testbeds ~\cite{mokhtarian2024survey, stager2018scaled, chalaki2021research} are often employed which offer controlled environments for implementing and testing algorithms. Scaled testbeds such as Duckietown \cite{paull2017duckietown} and the Robotarium \cite{pickem2017robotarium} have greatly advanced autonomy education and swarm coordination, but they are limited in scalability and do not support validation of the full autonomy stack of perception, planning, control, and communication required for AV research. 

Platforms tailored to CAVs, such as the CPM Lab \cite{scheffe2018scaled}, \textit{Cambridge Minicar} \cite{hyldmar2019fleet} and scaled smart city testbeds \cite{stager2018scaled}, enable evaluation of coordination and mixed-traffic interactions, while sensor-focused testbeds \cite{vargas2019development} emphasize the robustness of perception under adverse conditions. Despite their utility, these systems typically lack tightly integrated simulation environments which provide scalable virtual environments with diverse weather, lighting, and traffic conditions or scaling experiments beyond the physical limitations of available hardware. Such testbeds are also constrained by the number of robots that can be accommodated. Recent sim-to-real approaches \cite{abouchakra2025realissim} attempt to bridge this gap via digital twins, yet they remain primarily confined to robotic manipulation tasks. Finally, many of these testbeds are not publicly accessible, limiting broader support for AV/CAV research.

In contrast, our mixed-reality Hardware-In-the-Loop (HIL) smart city testbed unifies simulation and real robots, supports multimodal sensing and V2X communication, and enables research across perception, planning, control, and traffic network control and optimization. It combines the scenario diversity of simulation with the realism of physical hardware. Additionally, in our testbed, physical and simulated robots operate in a shared environment, allowing
us to test algorithms with a large number of agents. Our goal is to make the testbed \emph{open and remote-accessible following the review process to facilitate safe AV/CAV research particularly towards bridging the gap between simulation and hardware}. We present the primary contributions of our paper below:
\begin{itemize}   
    \item A mixed-reality HIL testbed that unifies a CARLA-based virtual simulator with physical mobile robots to bridge the gap between simulation and deployment. Our testbed provides several key features: scalability through mixed-reality integration, multimodal sensing (LiDAR, Cameras, IMU, etc.), real-time communication, and systematic data acquisition for learning-based methods. These capabilities support all aspects of AV research, including communication, perception, planning, and control, as well as research on multi-agent CAV systems.

    \item As a second major contribution, we propose and implement a novel end-to-end online self-supervised learning control framework for multi-agent systems including CAVs using Control Barrier Functions (CBF).
    The controller is implemented in a decentralized manner and validated within our mixed-reality testbed, showcasing the capabilities of the testbed and its ability to bridge the Sim2Real gap.
    The use of CBFs guarantees satisfaction of all safety constraints during both training and deployment. With the public release of the testbed, this framework enables the safe integration and evaluation of state-of-the-art (SOTA) autonomous vehicle (AV) algorithms by enforcing safety using our CBF controller.
\end{itemize}

The paper is organized as follows. 
%In section~\ref{Related_works}, we review related works in simulation platforms, HIL testbeds, and learning-based control for autonomous systems. 
%Section~\ref{premilinaries} presents the preliminaries for this paper. 
Section~\ref{Tetbed_Design} details 
the design of our HIL testbed. Next, section ~\ref{case_study} presents the proposed end-to-end safety-guaranteed framework and its integration with the CARLA simulator. The experimental results are presented and discussed in Section~\ref{exp_results}, demonstrating and validating the capabilities of our testbed. Finally, Section~\ref{sec:conclusion} concludes the paper and outlines future directions.

\section{Testbed Design}
\label{Tetbed_Design}
At first, we outline the key design considerations, followed by a detailed description of the implementation.
%In this section, we present the details of the testbed whose overall architecture is shown in figure \ref{fig:testbed} along with the end-to-end algorithm for safe planning and control of the autonomous vehicles.
 % \textcolor{red}{Testbed description, ros communication details, CARLA server, projection and teleport, decentralized e2e control, CARLA client, data collection and learning}
\begin{figure*}[h]
    \centering  \includegraphics[width=1\linewidth]{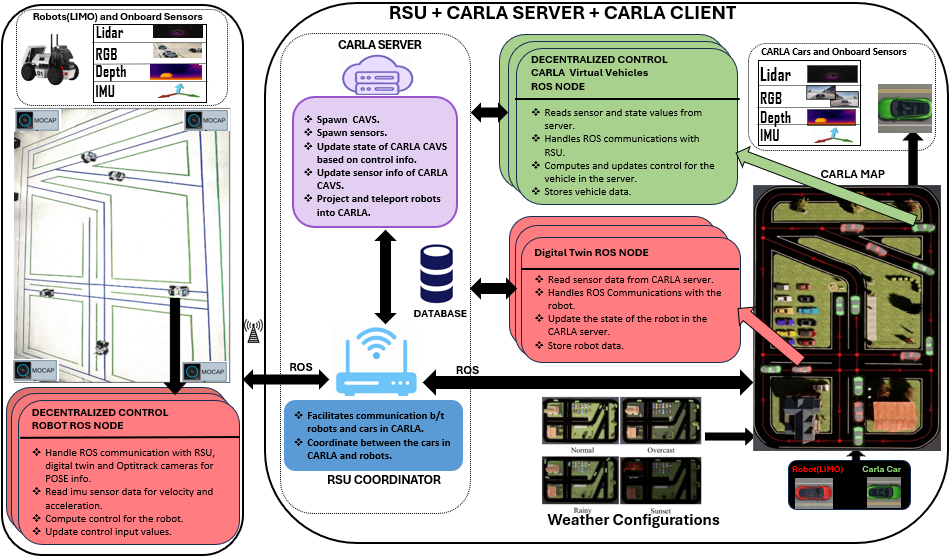}
    \caption{Overview of the mixed-reality HIL testbed using the CARLA simulator and physical mobile robots. The testbed comprises a physical arena with robots and a virtual simulator (CARLA) that spawns virtual agents and projects/“teleports” the physical robots into the digital scene to create mixed-reality simulations. A centralized roadside unit (RSU) facilitates V2I communication and coordinates traffic in the network. CARLA supports high-fidelity, multimodal sensor simulation (e.g., LiDAR, RGB cameras, depth cameras, and IMUs), enabling the physical robots to be augmented with photorealistic, real-world-like perception during mixed-reality experiments.}
    \label{fig:testbed}
\end{figure*}

\begin{figure}
    \centering
    \includegraphics[width=1\linewidth]{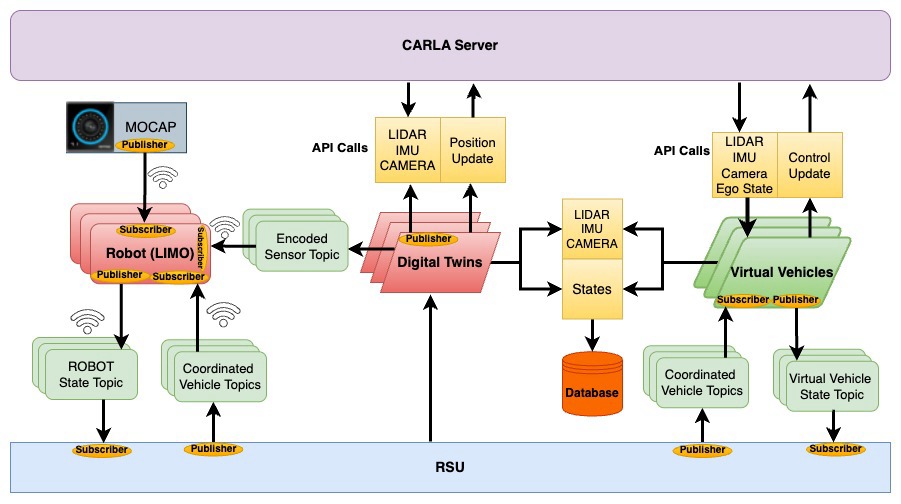}
    \caption{Data flow of the mixed-reality testbed.}
    \label{fig:flowchart}
\end{figure}

\subsection{Design Considerations}
\textbf{Mixed-Reality Simulation.}
We require combining the simulator's scalable, diverse scenario generation capability (e.g., CARLA) with physical robots to capture real-world uncertainties; this allows bridging the sim-to-real gap in autonomous driving (AD) research. 

\textbf{Multimodal Sensing.}
We integrate various sensor modalities including cameras, LiDAR, and radar to reflect modern AV perception stacks facilitating SOTA research.

\textbf{Real-Time Communication.}
Our testbed facilitates V2V and V2I communication required for research on traffic network optimization and the challenges associated with AV communication network technologies.

\textbf{Data Acquisition and Storage.}
Our design facilitates the storage of both multimodal sensor data as well as trajectory data to facilitate research on SOTA algorithms for AVs/CAVs.

\textbf{Motion planning and control (single and multi-agent systems).}  
We design our testbed infrastructure to support research and evaluation of both centralized and decentralized coordination/planning algorithms for AVs and CAVs.

\textbf{Scalability.} As mentioned, we aim to facilitate all aspects of AV research, including multi-agent CAV systems. To allow scalability, we adopt a mixed-reality testbed that scales efficiently by incorporating both virtual agents and physical ones.

\subsection{Testbed Implementation}
Guided by the design considerations outlined above, we have implemented our testbed using the architecture illustrated in Fig. \ref{fig:testbed}.

\textbf{Map Design.} We implemented our mixed-reality testbed  using the CARLA simulator as shown in Fig.~\ref{fig:testbed}. The smart city environment in CARLA was designed using custom scenes in MathWorks RoadRunner \cite{MathWorksRoadRunner}. It comprises of a diverse set of road network configurations such as merging lanes, intersections, and highway arteries — representative of typical urban traffic environments. The road network including lanes, intersections, and curbs was drafted in RoadRunner and enriched with markings, signals, landscaping, and buildings. In contrast to CARLA’s preconfigured maps, the custom map was scaled to align with a physical testbed layout as illustrated in Fig. \ref{fig:testbed}. The scene was exported as OpenDRIVE (.xodr) with Unreal Engine assets using RoadRunner's CARLA plug-in, and then converted with CARLA's Python tools into a map (\texttt{*.pak}) containing spawn points, waypoints and traffic scenarios. This workflow also supports real-world replication by importing GIS and elevation data (e.g., OpenStreetMap). 

\textbf{Mixed Reality Simulation.} Our testbed consists of three main components: a Road-Side coordinator Unit (RSU), the physical testbed with robots, and the CARLA simulator. The simulator and the RSU are hosted on the same PC, as illustrated in Fig.~\ref{fig:testbed}. We employ AgileX Limo robots~\cite{AgilexLimoPro} featuring Ackermann steering, making them suitable for autonomous vehicle (AV) research. CARLA operates under a client–server architecture, within which we instantiate two types of clients: (i) \textit{virtual vehicles} (shown in green on the right side of Fig.~\ref{fig:testbed}), and (ii) \textit{digital twins} (shown in red in Fig. \ref{fig:testbed}), which are virtual projections of the physical robots in the CARLA environment. The digital twins are virtual vehicles with their internal physics disabled in CARLA and states continuously synchronized with the real robots. This design enables a mixed-reality simulation that seamlessly integrates physical robots with virtual vehicles.

\textbf{Multimodal Sensing.} The AgileX Limo robots are equipped with LiDAR, RGB–Depth cameras, and IMU sensors. The CARLA simulator supports a wide range of sensing modalities, including LiDAR, RGB–Depth cameras, IMU, and radar. Compared to the physical platform, CARLA provides richer and more diverse visual scenes, making it particularly suitable for vision-based AD research. By combining CARLA’s virtual sensors with the planning and control stack running on physical robots, our framework enables realistic validation of full autonomy pipelines within a controlled laboratory environment. For the digital twins, we instantiate \textit{exteroceptive} sensors—LiDAR and cameras—to perceive the surrounding environment for planning and decision-making, while \textit{proprioceptive} sensors, such as the IMU, are used to estimate the robots’ internal states. Both the digital twins and the virtual vehicles access their respective sensor data via API calls to the CARLA server, as illustrated in Fig.~\ref{fig:flowchart}.

\textbf{Real-time communication.} Traffic coordination between physical robots and CARLA-simulated vehicles is managed by the RSU through the ROS framework. Our testbed supports both ROS 1 and ROS 2, enabled by a ROS 1–ROS 2 bridge for seamless interoperability. Vehicle-to-Infrastructure (V2I) communication is implemented via ROS topics, allowing both the physical robots and virtual vehicles to exchange information with the RSU, as illustrated in Figs. 1 and 2 respectively. In addition, MOtion CAPture (MOCAP) cameras provide real-time position and attitude estimates of the robots, which are published using ROS topics for robots to subscribe to them. The V2I messaging infrastructure operates over the lab’s private LAN, where the MOCAP workstation and the CARLA/RSU computer share the same routable subnet on a dedicated Wi-Fi network. This setup ensures low-latency, bidirectional ROS communication essential for real-time coordination and control.

\textbf{Scalability and Multi-Agent Simulation.} The RSU coordinates the robots, their digital twins, and virtual vehicles to enable cooperative multi-agent maneuvers. Each robot runs an onboard ROS node, as shown in Fig.~\ref{fig:testbed}. Similarly, each virtual vehicle and digital twin runs an individual ROS node on the same PC that hosts the RSU and the CARLA server. Both the physical robots and the CARLA agents continuously publish their state information—such as position, velocity, and attitude—to the RSU. Based on the coordination policy, the RSU disseminates relevant information about surrounding agents to each robot and virtual vehicle. The robots also receive exteroceptive sensor data from their corresponding digital twins through the RSU. By using the exteroceptive and onboard sensor data with the information received from the RSU, each robot computes and updates its control commands locally, as illustrated in Fig.~\ref{fig:flowchart}.

Meanwhile, each virtual vehicle uses their simulated sensor data from the CARLA server, combined with the RSU messages, computes and sends control inputs back to the CARLA environment. The digital twins subscribe to the RSU to receive their corresponding robots’ state information and update their poses in CARLA server which suppresses their physical
dynamics and continuously teleports them to synchronize
with the robots motion. This design ensures synchronization between the physical and virtual domains, allowing the robots and virtual vehicles to coexist in a mixed-reality environment.

\textbf{Data Acquisition and Storage.} 
In our testbed, robot data are stored by their corresponding digital twin ROS nodes, while CARLA vehicle data are stored by their respective ROS nodes in the database, as illustrated in Fig.~\ref{fig:flowchart}. 
Each digital twin and virtual agent logs its state information along with the states of cooperating vehicles received via the RSU as time-series data in \texttt{.pkl} format. Besides that, each agent stores their sensors data including image, LiDAR and IMU data. Images are saved in \texttt{.jpeg} format, LiDAR data are projected using the cylindrical projection method of \cite{chen_multi-view_2017} and stored in \texttt{.ply} format, and IMU data are stored in \texttt{.pkl} format. The decentralized architecture also allows additional fields to be incorporated into the agents' data structures to log further information as per necessity. 

\textbf{Web based Remote Access.} We have created a web application to allow remote access to the testbed. The interface allows users to configure the runs by (i) selecting the number of robots and CARLA virtual vehicles, (ii) specifying the origin, destination, and paths for each agent, and (iii) uploading the algorithm to run onboard both the robots and virtual vehicles. The web-interface allows for remote visualization of the mixed-reality simulation using the CARLA simulation window and the lab cameras. After each run, the interface enables downloading the logged data—including sensor measurements and agent trajectories—from the database. 
% Robots can use their own sensors or subscribe to sensor data generated by their digital twins in CARLA, giving them richer perception and awareness of virtual vehicles for planning. Both digital twins and virtual vehicles access sensor information via API calls to CARLA server. With centralized coordination, decentralized controllers, and CARLA’s server–client architecture, the system can easily scale to much larger numbers of vehicles. Communication is fully asynchronous, with no enforced synchronization or any need for global clock, further supporting scalability and flexibility. 

\section{Perception, Planning $\&$ Control Framework}
\label{case_study}
\begin{figure*}[h]
    \centering
    \includegraphics[width=1\linewidth]{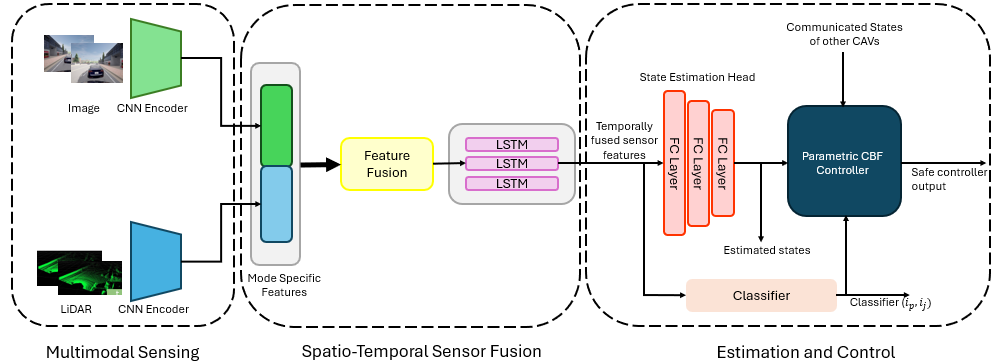}
    \caption{End-to-end framework comprising of multimodal sensing, sensor fusion, state estimation and CBF-based controller.}
    \label{fig:e2e_planning_control}
\end{figure*}
In this section, we present the decentralized end-to-end control pipeline comprising of perception, sensor fusion, state estimation and CBF-based control. The CBF-based design ensures inherent safety during both the training and validation phases. \emph{It is noteworthy that the testbed can be run with any algorithm that can generate reference for our CBF controller to ensure safety during testing.}
%{\color{blue} Bring up again that this design ensures safety even during training.}
\subsection{Multi-Agent Planning and Control}
Firstly, we present the novel online learning-based multi-agent CBF based control framework that involves centralized coordination and decentralized control. We consider $N$ CAVs/agents with a corresponding index set defined as $\mathcal{N} = \{1,\dots,N\}$. As shown in Fig. \ref{fig:testbed}, the road network has a set of conflict points (red circles) where multiple roads meet, (i.e., lateral collision is possible); we call these \emph{Merging Points (MPs)}. Therefore, at any given time, only one vehicle can traverse any particular MP. The set of MPs is denoted by 
%$M \in \{1,2,\dots, 14\}$.
$\mathcal{M} = \{1,2,\dots, m\}$, where $m$ depends on the scene/traffic network layout.
The state space and action space of any agent $i$ are denoted by $\mathcal{X}_i \subset \mathbb{R}^n$ and $\mathcal{U}_i \subset \mathbb{R}^m$ respectively. Therefore, the joint state space of the agents is denoted by $\mathcal{X} = \prod \mathcal{X}_i$ and $\mathcal{U} = \prod\mathcal{U}_i$ respectively. Each agent is considered to have an observation space $\mathcal{O}_i$ which includes the information from the sensors: LiDAR, Camera and IMU sensors. We denote 
%the origin and destination corresponding to a trip take by
%any agent $i$ as $p_{i}^o$ and $p_{i}^f$ respectively and use
$T_i$ as time taken by CAV $i$ to arrive at its destination.
%along with a route $r_i$ which is the set of segments the CAV will take to reach goal. For example, one possible route can be ${}$
The state dynamics are defined as follows:
\begin{align*}
\dot{s}_i &= \frac{v_i \cos(\mu_i + \beta_i)}{1 - d_i \kappa(s_i)}, \ \dot{d}_i = v_i \sin(\mu_i + \beta_i) \\
\dot{\mu}_i &= \frac{v_i}{l_r} \sin \beta_i - \kappa(s_i) \cdot \frac{v_i \cos(\mu_i + \beta_i)}{1 - d_i \kappa(s_i)} \\
\dot{v}_i &= a_i, \ \dot{\delta}_i = \omega_i 
\end{align*}
where $s_i(t)$ is the distance traveled by agent $i$ along its trajectory, $d_i(t)$ the signed lateral deviation from the centerline (positive left, negative right), $\mu_i(t)$ the heading error, $v_i(t)$ the linear speed, and $\delta_i(t)$ the steering angle. The control inputs are the longitudinal acceleration $a_i$ and steering rate $\omega_i$, bounded as $\boldsymbol{u}_{i,\min}=[a_{i,\min},\omega_{i,\min}]$ and $\boldsymbol{u}_{i,\max}=[a_{i,\max},\omega_{i,\max}]$, giving $\mathcal{U}_i=\{\boldsymbol{u}_i\in\mathbb{R}^2:\boldsymbol{u}_{i,\min}\leq \boldsymbol{u}_i\leq \boldsymbol{u}_{i,\max}\}$ where the vector inequalities are applied component-wise. The slip angle is $\beta_i(t)=\arctan\!\left(\tfrac{l_r}{l_r+l_f}\tan\delta_i(t)\right)$, where $l_r,l_f$ are distances from the vehicle center to the rear and front axles, and $\kappa(s_i)$ denotes the curvature of the reference trajectory at position $s_i$.

We define the cost over a single time step
for agent $i$ as $l_i(\boldsymbol{x}_i,\boldsymbol{u}_i)$, reflecting its performance and goal attainment (energy, fuel, and travel time). In the cooperative setting, the total cost is $L(\boldsymbol{x},\boldsymbol{u})=\sum_{i\in\mathcal{N}} l_i(\boldsymbol{x}_i,\boldsymbol{u}_i)$. Each agent $i\in\mathcal{N}$ is further associated with functions $\{b_i^j(\boldsymbol{x}_i,\boldsymbol{x}_j)\}_{j\in\mathcal{N}_i(\boldsymbol{x})}$, where $\boldsymbol{x}_j\in\mathcal{X}\setminus\mathcal{X}_i$ and $\mathcal{N}_i(\boldsymbol{x})$ is a finite set depending on the global state $\mathcal{X}$ corresponding to safety specifications expressed formally as shown next.

\noindent\textit{\textbf{Road Boundaries:}} The road boundary constraint \( b_{\text{road}}(x_i) = d_{\text{max}} - |d_i| \) can be equivalently expressed using two inequalities:
\begin{equation}
\label{road_boundary}
    -d_{\text{max}} \leq d_i(t) \leq d_{\text{max}}, \quad \forall t \in [0, T_i]
\end{equation}
\textit{\textbf{Rear-End Safety Constraint}}: Let 
\( \Delta(\boldsymbol{x}_{i_p}, \boldsymbol{x}_i) \)
be a differentiable (up to the order of the least actuator/input) distance metric between agent \( i \) and its preceding vehicle \( i_p \).
This metric makes the constraints agnostic to the road layout. As an example, in a straight road, the constraint takes the commonly adopted simple form $s_{i_p}(t) - s_i(t) - \phi v_i(t) -\delta \ge 0$,
%{\color{blue} Add references to one of our merging papers and one of the intersection papers (maybe the one with the generalized framework) where this form was used} ,
where $\phi$ is a typical response time and $\delta$ is a minimum distance between the centers of $i$ and $i_p$.
Then, the rear end safety constraint for CAV \( i \) can be expressed as:
\begin{equation}
\label{rear_safety}
    \Delta(\boldsymbol{x}_{i_p}(t), \boldsymbol{x}_i(t)) \ge 0, \quad \forall t \in [0, T_i]
\end{equation}
\textit{\textbf{Safe Merging Constraint:}} Let 
\( \Delta(\boldsymbol{x}_j, \boldsymbol{x}_i) \)
be a differentiable distance metric (up to the order of the least actuator/input) between agent \( i \) and a preceding merging vehicle \( j \). Same as above, this is agnostic to road geometry and the constraint for CAV \( i \) at a merging point can be expressed as
%{\color{blue} Show the case of as traight road as done above, for completness} 
:
\begin{equation}
\label{safe_merging}
    \Delta(\boldsymbol{x}_j(t), \boldsymbol{x}_i(t)) \ge 0, \quad \forall t \in [t^j_i, t^{m}_i]
\end{equation}
where $t_i^j$ is the time the constraint is first imposed when it arrives at a merging point
and $t_i^{m}$ is the time it is removed.
For straight road it can be: $s_j(t)-s_i(t)-\phi v_i(t) - \delta \ge 0$.
%{\color{blue} This is unclear. I know what you are trying to say, but I am certain that no reader/reviewer will get it! Going back to the numerous papers we have written that use this constraint, the point is that it applies ONLY at the time $i$ arrives at its next MP. So I suggest you use that as the time when (3) must apply. What is tricky here is that we don't know in advance what that time is. However, it can be evaluated upon solving an OCP, as done later. I believe it suffices to just say that here and defer the evaluation of this time, say $t_i^m$, to the sequel.}

In our fully cooperative setting
containing CAVs, a roadside unit (RSU) coordinates the traffic. They communicate their states, origin and destination information to the RSU which uses this information to coordinate traffic by identifying the order CAVs traverse through each merging point in their path to their destination. 
 Observe that a CAV can have at most two merging constraints, which happens while traveling through an intersection. Therefore, there are at most three cooperating CAVs for each CAV $i$, i.e., $\max_t|\mathcal{N}_i(\boldsymbol{x})| = 3$: ($i,i_p$) for the rear end safety constraint \eqref{rear_safety}  and ($i,i_j$) 
for the safe merging constraint \eqref{safe_merging}). The coordinator communicates the information of these vehicles to CAV $i$. 
%{\color{blue} What you are trying to say here (with no need for extra notation) is that the coordinator can obtain state information for CAVs that affect $i$ and communicate that to $i$ (tat would be $i_p$ in (2) and $i_j$ in (3) if you wat to b precise). Please make sure to fix this!}
%{\color{blue} I don't see any immediate implication to justify "therefore"! This is a new statement about how to formulate an OCP. I would just write in a new paragraph:}
%We can now formulate the following decentralized Optimal Control Problem (OCP) with safety constraints for each $i$:
%the decentralized control problem for any agent $i$ can be formulated as an Optimal Control Problem (OCP) and mapped to a problem using CBFs as they are sufficient conditions as follows:
%\begin{align*}
%\min_{\boldsymbol{u}_i(\cdot), T_i} \quad & \int_0^{T_i} l_i(\boldsymbol{x}_i(t), \boldsymbol{u}_i(t)) \, dt \\
%\text{s.t.} \quad & \dot{\boldsymbol{x}}_i(t) = f_i(\boldsymbol{x}_i(t), \boldsymbol{u}_i(t)) \\
%& b_i^j(\boldsymbol{x}_i(t), \boldsymbol{x}_j(t)) \ge 0, \forall j \in \mathcal{N}_i(\boldsymbol{x}), \forall t \in [0, T_i] \\
%& \boldsymbol{x}_i(0) = \boldsymbol{x}_{i,0}, \quad \boldsymbol{x}_i(T_i) = \boldsymbol{x}_{i,T_i}  \boldsymbol{u}_i(t) \in \mathcal{U}_i, \quad \forall t \in [0, T_i] 
%\end{align*}
We leverage CBF theory \cite{xiao2023safe} to replace state constraints $b(\boldsymbol{x}) \ge 0$ with CBF constraints. 
{ This results in a trade-off between guaranteed safety and some performance loss.
}
%\begin{equation} \label{HOCBF}
%\sup_{\boldsymbol{u}\in \mathcal{U}} \big[L_fb(\boldsymbol{x})+L_gb(\boldsymbol{x})\boldsymbol{u}+\alpha_m(b(\boldsymbol{x}))\big] \ge 0,
%\end{equation}
%where $L_f$ and $L_g$ are Lie derivatives 
%of the function $b(\boldsymbol{x})$ along the dynamics $\dot{\boldsymbol{x}}=f(\boldsymbol{x})+g(\boldsymbol{x})\boldsymbol{u}$.''
%and $\alpha_m(\cdot)$ is a class-$K$ function \cite{xiao2023safe}.
%If $L_gb(\boldsymbol{x})=0$, the condition extends to High-Order CBFs (HOCBFs) by differentiating $b(\boldsymbol{x})$ up to order $m$ until $\boldsymbol{u}$ appears. Condition (\ref{HOCBF}) provides a sufficient, 
%therefore possibly conservative,
%safety guarantee. 
The OCP with CBF-based constraints for agent $i$ is as follows:
\begin{align}
\min_{\boldsymbol{u}_i(\cdot), T_i} \quad & \int_0^{T_i} l_i(\boldsymbol{x}_i(t), \boldsymbol{u}_i(t)) \, dt  \\
\text{s.t.} \quad & \dot{\boldsymbol{x}}_i(t) = f_i(\boldsymbol{x}_i(t), \boldsymbol{u}_i(t)) \nonumber \\
& L_f^mb_i^j(\boldsymbol{x})+L_gL_f^{m-1}b_i^j(\boldsymbol{x})\boldsymbol{u}+S(b_i^j(\boldsymbol{x}))  \nonumber \\
& + \alpha_m(\zeta_{m-1}(\boldsymbol{x})) \ge 0, \forall j \in \mathcal{N}_i(\boldsymbol{x}), \forall t \in [0, T_i] \nonumber \\
& \boldsymbol{x}_i(0) = \boldsymbol{x}_{i,0}, \boldsymbol{x}_i(T_i) = \boldsymbol{x}_{i,T_i}, \boldsymbol{u}_i(t) \in \mathcal{U}_i, \forall t \in [0, T_i]  \nonumber
\end{align}
where $T_i$ is a decision variable corresponding to the minimal time required to reach the terminal state $\boldsymbol{x}_i(T_i) = \boldsymbol{x}_{i,T_i}$, $L_fb$ and $L_gb$ are Lie derivatives 
of the function $b(\boldsymbol{x})$ along the vector field $f$ and $g$  respectively. Here we have used High Order CBFs (HOCBF) in which the term 
$S(b_i^j(\boldsymbol{x}))$ is the sum of lower than $m$ order Lie derivatives and 
$\zeta_{m}(\boldsymbol{x})$
is: 
\begin{equation} \label{eqn:zetafunctions}
\zeta_i(\bm x) = \dot \zeta_{i-1}(\bm x) \!+\! \alpha_i(\zeta_{i-1}(\bm x,)), ~~~ i\in\{1,\dots,m\},
\end{equation}
where $\zeta_0(\bm x) = b(\bm x)$.
The coordination policy which determines the index $j$ in (\ref{safe_merging}) can be implemented in several ways based on relative distance from the MP or the order of arrival of CAVs 
%in the incoming roads 
or based on the traffic density at the incoming roads, to name a few;
details may be found in \cite{xiao2023safe}.

The transformed OCP above benefits from the linear in the control structure of the safety constraints and is typically solved (see \cite{xiao2023safe}, \cite{Ames_01}) by discretizing time 
%{\color{blue} Bring up here the event-driven improvement which frees us from having to select a time step. You can probably do that in the form of a Remark. Also, cite our recent paper that does that.}
and solving a sequence of Quadratic Programs (QPs), one at each time step. We parameterize the QP to adapt the controller against uncertainties and for balancing the tradeoff between safety and conservativeness. In particular, at each time step CAV $i$ solves the following parametric CBF-QP to compute a safe control input:
\begin{align}
\boldsymbol{u}_i^*(\boldsymbol{x}_i,\tilde{\boldsymbol{x}}_i; \boldsymbol{\theta}_c) \label{eq:controller} &=  \\
\arg\min_{\boldsymbol{u}_i \in \mathcal{U}_i} \quad & ||\boldsymbol{u}_i - \boldsymbol{u}_i^{\text{ref}}(\boldsymbol{x}_i, \tilde{\boldsymbol{x}}_i; \boldsymbol{\theta}_u)||^2_{W(\boldsymbol{x}_i, \tilde{\boldsymbol{x}}_i;\boldsymbol{\theta}_W) } \nonumber \nonumber \\ 
\text{s.t.} \quad 
 L_f^m  & b_i^j(\boldsymbol{x}_i,  \boldsymbol{x}_j) 
+ L_g L_f^{m-1} b_i^j(\boldsymbol{x}_i, \boldsymbol{x}_j) \, \boldsymbol{u}_i + \nonumber \\ S(b_i^j & (\boldsymbol{x}_i, \boldsymbol{x}_j); \boldsymbol{\theta}_{i,j}) + \alpha_m(\zeta_{m-1}(\boldsymbol{x}_i, \boldsymbol{x}_j); \boldsymbol{\theta}_{i,j}) 
\geq 0, \nonumber \\ \forall j \in & \mathcal{N}_i(\boldsymbol{x}), \forall t \in [0,T_i] \nonumber
\end{align}
where $\tilde{\boldsymbol{x}}_i = [\boldsymbol{x}_{i_p}, \boldsymbol{x}_{j}], \ \forall j \in \mathcal{N}_i(\boldsymbol{x}) $  is the stacked vector of all the states of the CAVs that CAV $i$ cooperate with
and  \( \boldsymbol{u}_i^{\text{ref}}(\boldsymbol{x}_i,\tilde{\boldsymbol{x}}_i; \boldsymbol{\theta}_u) \in \mathcal{U}_i\) is a nominal state feedback control input computed as a parameterized function of \( \boldsymbol{\theta}_u\),
while the weight factor $W(\boldsymbol{x}_i, \tilde{\boldsymbol{x}}_i;\boldsymbol{\theta}_W)$
is parameterized by $\boldsymbol{\theta}_W$. Finally, the class-$K$ function 
$\alpha_m(\zeta_{m-1}(\boldsymbol{x}_i, \boldsymbol{x}_j); \boldsymbol{\theta}_{i,j})$
is also parameterized by $\theta_{i,j}$. Therefore, the control parameter vector is $\boldsymbol{\theta}_c = [\boldsymbol{\theta}_u, \boldsymbol{\theta}_W,\boldsymbol{\theta}_{i,1}, \dots, \boldsymbol{\theta}_{i,|\mathcal{N}_i|}]$. 
Next, we present an online self-supervised learning approach to learn these parameters. 

\subsubsection{Online learning of the Controller}
%As mentioned, we consider a cooperative setting where the RSU coordinates the traffic. 
We consider stationary (i.e., time-invariant, state-dependent) coordination policies and define the set of feasible policies for the given coordination scheme based on $\tilde{\boldsymbol{x}}_i$ as $\tilde{\Pi}_i$. Thereafter, we formulate the learning problem as follows:
\begin{align}
\label{online_loss}
J(\boldsymbol{\theta}_c) =& \min_{\boldsymbol{\theta}_c} \, \mathbb{E}_{\tilde{\pi}_i \in \tilde{\Pi}_i}[\mathbb{E}_{\boldsymbol{x}_i \sim d^{{u*}(\cdot;\boldsymbol{\theta}_c)}, \boldsymbol{\tilde{x}}_i \sim  d^{\tilde{\pi}}} [ w_ll_i\big(\boldsymbol{x}_i, \boldsymbol{u}^*(\boldsymbol{x}_i,\boldsymbol{\tilde{x}}_i\nonumber \\
& ;\boldsymbol{\theta}) \big)- \sum_{j \in \mathcal{N}_i(\boldsymbol{x})}w_{b}^j\min(0, b_i^j(\boldsymbol{x}_i,\boldsymbol{x}_j)] ] \\
= \min_{\boldsymbol{\theta}_c} \, & \mathbb{E}_{\tilde{\pi}_i \in \tilde{\Pi}_i}[\mathbb{E}_{\boldsymbol{x}_i \sim d^{{u*}(\cdot;\boldsymbol{\theta}_c)}, \boldsymbol{\tilde{x}}_i \sim  d^{\tilde{\pi}}}[\tilde{l}_i(\boldsymbol{x}_i,\tilde{\boldsymbol{x}}_i,\boldsymbol{u}^*(\boldsymbol{x}_i,\tilde{\boldsymbol{x}}_i,\boldsymbol{\theta}_c))]]  \nonumber
\end{align}
where $d^{\boldsymbol{u}^*(.;\boldsymbol{\theta}_c)}(\boldsymbol{x}_i)$ and  $d^{\tilde{\pi}}(\tilde{\boldsymbol{x}_i})$ are stationary state distributions of the ego CAV $i$ and CAVs in $\mathcal{N}_i(\boldsymbol{x}_i)$ respectively and $w_l, w_b^j\in\mathbb{R}_{>0}$ are constant weights.  We aim to learn the parameters that optimize the objective in \eqref{online_loss} as a function of the states of the ego vehicle and other vehicles, i.e.,  
$\boldsymbol{\theta}_c = \boldsymbol{\theta}_c(\boldsymbol{x}_i, \tilde{\boldsymbol{x}}_i; \phi)$,
where $\phi$ are the parameters of a NN.
% The gradient of the objective function with respect to the parameters can be computed using Karush–Kuhn–Tucker conditions 
%{\color{green} The Karush–Kuhn–Tucker conditions characterize the optimal solution, and by differentiating these optimality conditions we obtain the gradient of the objective function with respect to the parameters, as in \cite{amos2017optnet}.}
%{\color{blue} The gradient cannot be computed by the KKT conditions! The KKT conditions simply tell us when optimality is achived using the gradient which someone else has computed! Please reword.}
 Let $\boldsymbol{u}^\star$ be the optimal control and $\boldsymbol{\lambda}^\star$ the optimal dual variables of trainable HOCBF constraints. The gradients of the objective with respect to the QP parameters $\boldsymbol{\theta}_c$ can be computed following \cite{pmlr-v70-amos17a} using the chain rule:
 \begin{equation}
 \label{grad_QP}
 \begin{aligned}
 \nabla_H\tilde{l} &=\frac12\!\bigl(d_{\boldsymbol{u}}\boldsymbol{u}^{\star\!\top}
                                 +\boldsymbol{u}^\star d_{\boldsymbol{u}}^{\!\top}\bigr) \quad
 \nabla_F\tilde{l} = d_{\boldsymbol{u}},\\
 \nabla_G\tilde{l} &= D(\boldsymbol{\lambda}^\star)
                       \bigl(d_{\boldsymbol{\lambda}}\boldsymbol{u}^{\star\!\top}
                           +\boldsymbol{\lambda}^\star d_{\boldsymbol{u}}^{\!\top}\bigr), \quad
 \nabla_h\tilde{l} = -\,D(\boldsymbol{\lambda}^\star)d_{\boldsymbol{\lambda}} \\
 \nabla_{W}\,\tilde{l} &= 2\,\nabla_{H}\tilde{l}\;-\;2\,\nabla_{F}\tilde{l}\,\boldsymbol{u}_{\mathrm{ref}}^T, \quad \nabla_{\boldsymbol{u}_{\mathrm{ref}}}\tilde{l} = -\,2\,W^{\!\top}\,\nabla_{F}\tilde{l}
 \end{aligned}
 \end{equation}
 Setting $G$ and $h$ as concatenated vectors with $j \in \mathcal{N}_i(\boldsymbol{x})$:

 \begin{equation}     
\begin{aligned}
 &G^{j}_{\!b} =-\,L_g L_f^{m-1} b_i^{\,j}(\boldsymbol{x}_i,\boldsymbol{x}_j), \quad \\
 &h^{j}_{b}=L_f^{m} b_i^j(\boldsymbol{x}_i,\boldsymbol{x}_j)           +\Omega\bigl(b_i^j(\boldsymbol{x}_i,\boldsymbol{x}_j);\boldsymbol{\phi}_b^{i,j}\bigr)   \\ & +\alpha_m\!\bigl(\zeta_{m-1}(\boldsymbol{x}_i,\boldsymbol{x}_j);\boldsymbol{\phi}_b^{i,j}\bigr) \nonumber
 \end{aligned}
 \end{equation}
 and using the diagonalizing operator $D(\mathbf{v})=\operatorname{diag}(\mathbf{v})$,
 $d_{\boldsymbol{u}}$ and $d_{\boldsymbol{\lambda}}$ can be computed by solving the linear  system of equations:
 \begin{equation}
 \label{kkt_system}
 \begin{bmatrix}
 H & G^\top D(\boldsymbol{\lambda}^\star)\\[3pt]
 G & D\!\bigl(G\boldsymbol{u}^\star-h\bigr)
 \end{bmatrix}
\!
 \begin{bmatrix}
 d_{\boldsymbol{u}}\\[2pt] d_{\boldsymbol{\lambda}}
 \end{bmatrix}
 =
 \begin{bmatrix}
 \mathbf{1}_{m}\\[2pt] 0
 \end{bmatrix}
 \end{equation}
The controller parameters can be learned online by collecting driving data in a self-supervised manner without requiring a reward signal/labeled data. The corresponding learning algorithm is provided in Algorithm \ref{alg:online_learning}.
%{\color{blue} Again, I would stress safety during training!}
where it is noteworthy that the CBF constraints are enforced throughout, {hence the training process is guaranteed to be safe.} 

\begin{algorithm}[h]
    \caption{Self-Supervised Online Learning Framework}
\textbf{Input:} Initial controller default parameters $\boldsymbol{\theta}_c$  \\
\textbf{Output:} Learned optimal controller parameters $\boldsymbol{\theta}^*_c$ \\
\textbf{Data Collection:}
Run episode by using $\boldsymbol{u}^*(\boldsymbol{x}_i,\tilde{\boldsymbol{x}}_i)$ as action for the ego CAV $i$ and sample actions for other CAVs using some policy $\boldsymbol{\tilde{u}_i} \sim \tilde{\pi}_i$ and collect data $\mathcal{D} =  \bigl\{\,\bigl(\boldsymbol{x}_{i}^{(t)},\;\tilde{\boldsymbol{x}}_{i}^{(t)}\bigr)\bigr\}_{t=1}^{N}.
$
\label{alg:online_learning}
\begin{algorithmic}
    \item Initialize reference control, objective weights, and hyperparameters pertaining to CBF constraints by initializing parameter $\boldsymbol{\theta}_c$, the learning rate $\gamma$, and the number of Epochs, 
    % \While{Epochs}
    \While{$k \le K$}
    \State Sample a mini-batch of size $B$ from $\mathcal{D}$.
    \State Estimate the objective gradient:
    \[
      \nabla_{\boldsymbol{\theta}_c}J(\boldsymbol{\theta}_c) \gets
      \frac{1}{B}\sum_{b=1}^{B}
      \nabla_{\boldsymbol{\theta}_c}\tilde{l}_i\!\Bigl(
          \boldsymbol{x}_{i}^{(b)},
          u^\ast(\boldsymbol{x}_{i}^{(b)},\tilde{\boldsymbol{x}}_{i}^{(b)};\theta)
      \Bigr)
    \]
    \State Update parameters: $ \boldsymbol{\theta}_c \gets
      \boldsymbol{\theta}_c - \gamma \,\nabla_{\boldsymbol{\theta}_c}J(\boldsymbol{\theta}_c)
    $
    \State $k \gets k+1$
\EndWhile
\end{algorithmic}
\end{algorithm}

\subsection{Sensor-Fusion based State Estimation}
Next, we present the details of the multimodal sensor fusion algorithm used for state estimation. Many modern AD algorithms are implemented end-to-end using multimodal sensor data. Our pipeline thus serves as a means to demonstrate and validate the capability of our testbed to facilitate SOTA AD research. We use the sensor fusion pipeline proposed in \cite{chen2019selective} to learn the state estimates for the controller from the sensor observations. The overall pipeline is illustrated in Fig. \ref{fig:e2e_planning_control}. The fusion module provides state estimates for both the ego vehicle and the states of the nearby vehicle. Specifically, we estimate the state of the nearby vehicle that is in our line of motion. The states of the ego vehicle include: the lateral deviation $\hat{d}_i$, and velocity $\hat{v}_i$. The estimated states of the nearby vehicle include its velocity $\hat{v}_{i_p}$ and the relative distance to it from the ego vehicle $\hat{s}_{i_p} - \hat{s}_i$. In addition, we train a classifier to determine/classify if the vehicle is in the path of the motion of the ego vehicle. Overall, the pipeline contains RGB and LiDAR encoders with a fusion network followed by the estimation head. Subsequently, the estimated states are used to incorporate additional constraints on the controller namely the rear-end constraint \eqref{rear_safety} involving the vehicle in the line of motion and lane keeping constraint \eqref{safe_merging} based on the lateral deviation. The end-to-end CBF controller acts as a safety layer for arbitrary AV/CAV algorithms by taking the control input generated by the algorithm as $u_{ref}$ in \eqref{eq:controller} and projecting it onto the admissible control set defined by the CBF constraints.

\section{Experimental Results}
\label{exp_results}
%\subsection{Implementation Details}
%\subsection{Training and Evaluation Results}

%\textbf{Differentiable QP training}

%\textbf{Sensor fusion training}

This section presents the experimental results that validate the functionalities of the testbed. 

\subsection{Online Self-Supervised Learning Framework } 
\begin{figure}[h]
    \centering
    \includegraphics[width=1\linewidth]{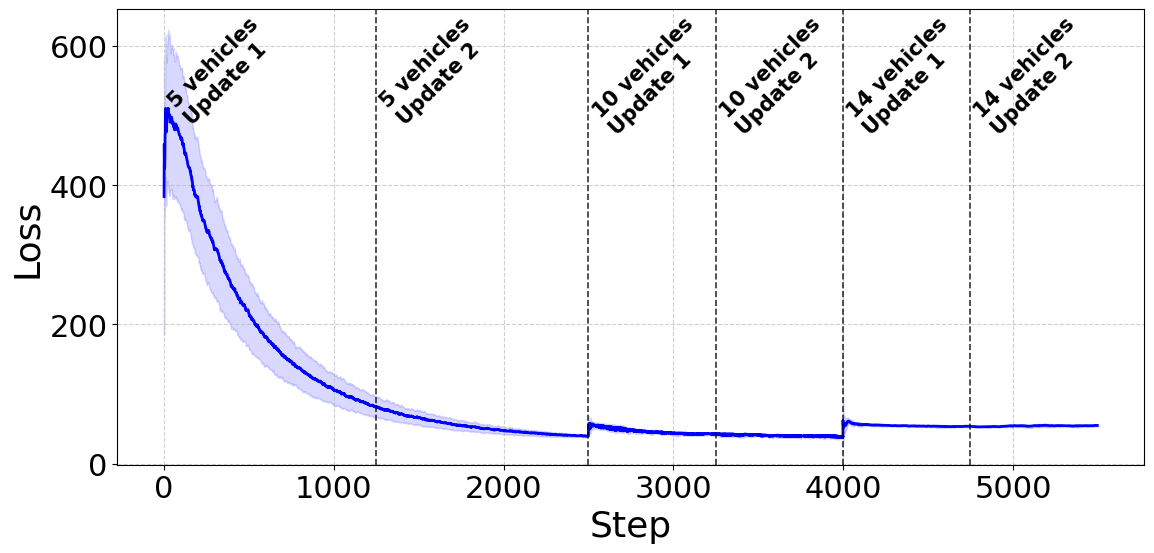}
    \caption{Online self-supervised algorithm training result in a multi-agent setting. The agents are considered to be homogeneous. Data from all agents are collected and used to train and update the controller across all agents.}
    \label{fig:dQP_result}
\end{figure}

At first, we present the results of our online self-supervised learning framework. %and assess its effectiveness in bridging Sim2Real gap. 
Training is set up to optimize the online control loss corresponding to the optimization problem in \eqref{online_loss}. The training phase is initiated using a stable nominal controller that commands a fixed reference acceleration in \eqref{eq:controller}. Note that, although no explicit velocity constraints are imposed at the high level, they are implicitly enforced by the low-level control loops of the physical robots and by the native vehicle models. We first train the controller in simulation using CARLA with virtual vehicles and subsequently fine tune it with robot data collected using the testbed. To improve training performance, we adopt a curriculum learning approach similar to \cite{ahmad2025hierarchical} varying traffic densities. For each curriculum, data from all vehicles -- assumed homogeneous -- is aggregated and used to train and update the controller. We run three different curricula comprising of five, ten and twelve vehicles, and use the collected data to train and update the controller.  This process is repeated twice per curriculum in an on-policy fashion. Fig. ~\ref{fig:dQP_result} shows the training results of our proposed algorithm. 
Each curriculum uses the learned model from the previous curriculum i.e., traffic density.
As can be seen, the training loss in equation \eqref{online_loss} converges within a few iterations, requiring only two policy updates for each curriculum, even with a small number of CAVs. The discrete jumps observed in the figure are caused due to jump in traffic density. However, within each curriculum, the loss decreases consistently within a few iterations and the policy converges after two updates.
%{\color{blue} Is it acceleration or speed? I still have this question since a positive acceleration will cause a vehicle to go unstable or soon hit its $v_{\max}$! If indeed the acceleration was fixed, please explain the rationale. Is it that you just ensured that $v_{\max}$ was never exceeded? Even so, what is the rationale for fixing acceleration ad not speed?}
% \begin{figure}
%     \centering
%     \includegraphics[width=1\linewidth]{training_loss_over_steps4.png}
%     \caption{Enter Caption}
%     \label{fig:placeholder}
% \end{figure}
% \begin{figure}
%     \centering
%     \includegraphics[width=1\linewidth]{sim_robot_training_result.png}
%     \caption{Enter Caption}
%     \label{fig:placeholder}
% \end{figure}
\subsection{Sensor-Fusion based State Estimation}
\begin{figure}[h]
    \centering
    \includegraphics[width=0.8\linewidth]{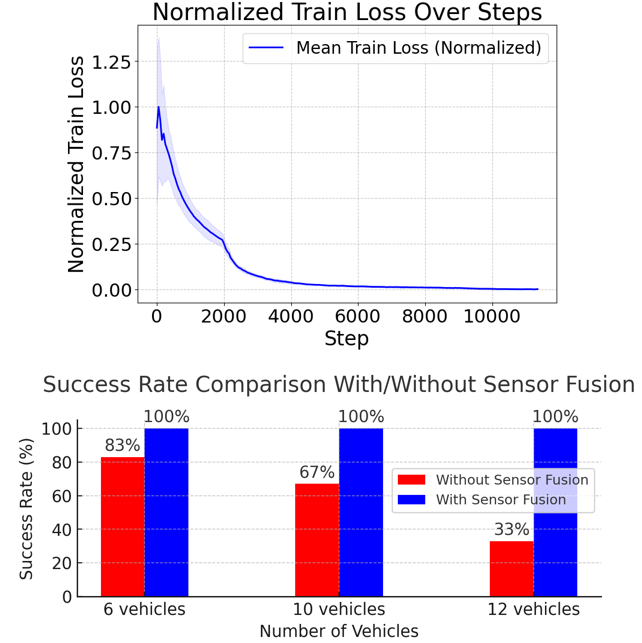}
    \caption{Training and validation results for Sensor Fusion.}
    \label{fig:e2e_results}
\end{figure}

Next, we present the training and validation results for the sensor fusion network. We collect data using the learned controller in simulation in CARLA and train the fusion pipeline by freezing the controller to minimize the state estimation loss and the classifier loss in Fig. \ref{fig:e2e_planning_control}. The training result shown in Fig.~\ref{fig:e2e_results} demonstrates that the loss is successfully minimized within few iterations. To validate the perception system, we introduced synthetic communication noise by uniformly sampling perturbations in the ranges [0, 5]m for GPS measurements and [0, 2]rad/s for IMU velocity readings. Experiments were carried out across multiple traffic densities and the results are presented in Fig.~\ref{fig:e2e_results}. As observed, this led to an increase in collisions and a corresponding reduction in success rate, an issue that was mitigated through the incorporation of the perception component, thereby validating its effectiveness.

These results underscore an important capability of our mixed-reality testbed: the ability to simulate photorealistic environmental conditions for the development and validation of perception to planning algorithms using real hardware. A further outcome of this setup is the availability of paired real actions and virtual sensor data, which can be leveraged to develop vision–action models that interact with a rich virtual world while remaining grounded in physical execution.

% To validate the perception system, we introduced synthetic communication noise by uniformly sampling perturbations in the ranges [0, 5] m for GPS measurements and [0, 2] rad/s for IMU velocity readings. Experiments were carried out across multiple traffic densities and the results are presented in figure \ref{fig:e2e_results}. As shown in figure \ref{fig:e2e_results} , this led to an increase in collision reducing the success rate, which was eliminated with the perception network thus validating its performance. These results underscore two important capabilities of our hardware-in-the-loop testbed: (i) the ability to simulate photorealistic environmental conditions for the development and validation of perception to planning algorithms using real hardware, and (ii) seamless integration of realistic wireless communication models to support the research and development of autonomous-vehicle networking protocols and algorithms.

\subsection{End-to-End Controller}
\begin{figure}[h]
    \centering
    \includegraphics[width=1\linewidth]{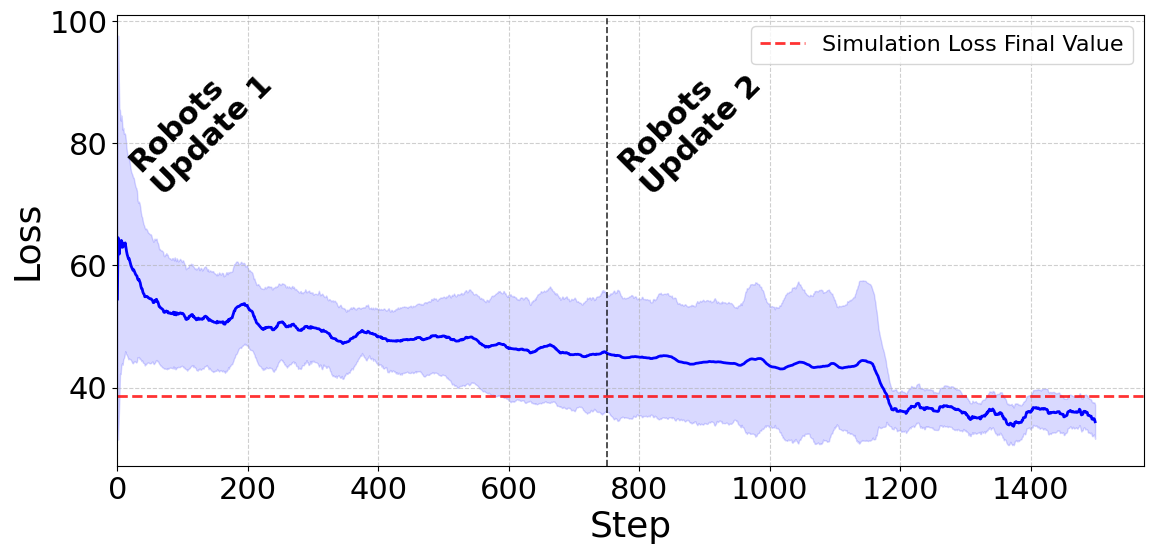}
    \caption{Plot of training loss with robot data for 5 seeds. With fine-tuning, there is virtually no gap between simulation loss (red line) and the loss with the robots.}
    \label{fig:simrealgap}
\end{figure}

Finally, we present the result of the end-to-end controller with the sensor fusion network and the online learned controller. The process leverages all the core capabilities of the testbed, and, therefore, serves to validate our design and implementation. In order to bridge the sim to reality gap, we implement the end-to-end controller on the robots and fine-tune it w.r.t the online control loss given by \eqref{online_loss} using the collected real-world state–action trajectories. This fine-tuning step is intended to compensate for modeling errors and unmodeled dynamics present in the real system. We collected real-world data by deploying the trained controller from simulation data on a fleet of 12 vehicles, including 5 physical robots. We updated the policy twice whereby for each update we ran the CAVs (robots and virtual vehicles) for only one episode.
%\wenchao{fine-tune with what? more details are needed here}
Presented in Fig.~\ref{fig:simrealgap} is the plot of the loss in \eqref{online_loss}. The red line corresponds to the training loss achieved upon convergence in simulation.  It can be noticed that fine-tuning decreases the loss and the policy achieves similar loss to the simulation loss within two policy updates and few ($\sim $ 1500) training iterations. This illustrates the ability of our testbed to allow for bridging the sim-to-real gap through fine-tuning. %Additional details of the training particularly the hyperparemters can be found in the added supplementary materials.
%\wenchao{Can we have direct measurements of the sim-to-real gap?}
Next, to evaluate the effectiveness of our method, we compare it against baselines using three metrics: 
\begin{itemize}
    \item \emph{Success rate}: the proportion of vehicles that reach their destination without collisions.
    \item \emph{Success-rate-weighted fuel consumption}: the average fuel usage of successful agents divided by the success rate.
    \item \emph{Success-rate-weighted travel time}: the average travel time of successful agents divided by the success rate.
\end{itemize}
% The first is the success rate, defined as the proportion of vehicles that reach their destination without collisions. The second and third are success-rate-weighted fuel consumption and travel time, which capture the average performance of successful agents while accounting for success rate. 
Weighting ensures fair comparisons and avoids misleading results, such as when a single vehicle completes the task while the others crash, which could otherwise suggest unrealistically low energy and fuel usage. To construct the baseline, we tune the controller by selecting the nominal control input $\boldsymbol{u}^{\text{ref}}$ and the CBF constraints under conservative and aggressive modes. The distinction lies in the magnitude of $\boldsymbol{u}^{\text{ref}}$ and the rate/limits on the derivatives of the CBF constraints by adjusting the class-$K$ function parameters. The conservative mode uses smaller acceleration and sublinear class-$K$ function for the CBF constraints, while the aggressive mode employs larger reference acceleration and superlinear class-$K$ functions. 
In addition, we include the 
\emph{Intelligent Driver Model (IDM)} baseline~\cite{treiber2000congested}
a widely used microscopic traffic model where the longitudinal acceleration is defined as

\begin{equation}
    \dot{v} = a \left(1 - \left(\frac{v}{v_0}\right)^\delta - \left(\frac{s^*(v,\Delta v)}{s}\right)^2\right),
\end{equation}
with desired velocity $v_0$, maximum acceleration $a$, and safe distance function $s^*(v,\Delta v)$. 

We evaluate our end-to-end framework against the aforementioned baseline approaches. 
The results of this evaluation are summarized in Table~\ref{res:baseline_comparison_2}.  We used a range of traffic scenarios with varying densities, in which more than $50\%$ of the agents were physical robots. Data were collected from both CARLA vehicles and physical robots to compute all evaluation metrics. As shown, our method achieves superior performance across all metrics and traffic densities when compared to the manually tuned baseline controller and IDM, validating the effectiveness of our approach. Notably, our trained controller consistently outperformed all baseline methods across multiple metrics and traffic conditions. These results demonstrate both the effectiveness of the proposed control framework and the fidelity of the HIL testbed in enabling reliable sim-to-real transfer. 
\begin{table}[h]
    \centering
    \renewcommand{\arraystretch}{1.25}
    \caption{\textit{Summary of results for various controller parameter choices with hardware-in-the-loop.}}
    {\scriptsize
  \begin{tabular}{|c|c|ccc|}
        \hline
        \textbf{Cars} & \textbf{Method} & Success & Fuel & Time \\
        \hline
        \multirow{4}{*}{5} 
        & Cons. & $0.67_{\pm 0.25}$ & $621.07_{\pm 150.08}$ & $50.97_{\pm 14.90}$ \\
        & Aggr. & $0.70_{\pm 0.10}$ & $562.03_{\pm 1.75}$   & $36.23_{\pm 1.44}$ \\
        & IDM   & $0.67_{\pm 0.09}$ & $588.88_{\pm 48.71}$  & $34.8_{\pm 0.92}$ \\
        & \textbf{Ours} & $\mathbf{1.00_{\pm 0.00}}$ & $\mathbf{409.80_{\pm 24.62}}$ & $\mathbf{37.60_{\pm 3.34}}$ \\
        \hline
        \multirow{4}{*}{10} 
        & Cons. & $0.77_{\pm 0.21}$ & $535.09_{\pm 343.93}$ & $43.86_{\pm 14.15}$ \\
        & Aggr. & $0.63_{\pm 0.12}$ & $543.31_{\pm 38.80}$  & $52.90_{\pm 18.75}$ \\
        & IDM   & $0.73_{\pm 0.11}$ & $485.82_{\pm 36.70}$  & $45.32_{\pm 1.06}$ \\
        & \textbf{Ours} & $\mathbf{1.00_{\pm 0.00}}$ & $\mathbf{378.39_{\pm 35.98}}$ & $\mathbf{39.52_{\pm 1.57}}$ \\
        \hline
        \multirow{4}{*}{12} 
        & Cons. & $0.585_{\pm 0.120}$ & $536.88_{\pm 113.58}$ & $67.21_{\pm 0.47}$ \\
        & Aggr. & $0.50_{\pm 0.00}$   & $636.47_{\pm 45.35}$  & $89.83_{\pm 0.32}$ \\
        & IDM   & $0.53_{\pm 0.03}$   & $533.46_{\pm 37.33}$  & $57.17_{\pm 0.75}$ \\
        & \textbf{Ours} & $\mathbf{1.00_{\pm 0.00}}$ & $\mathbf{392.60_{\pm 1.07}}$ & $\mathbf{44.29_{\pm 0.03}}$ \\
        \hline
    \end{tabular}
    }
    \label{res:baseline_comparison_2}
\end{table}

\begin{figure}[h]
    \centering
    \includegraphics[width=1\linewidth]{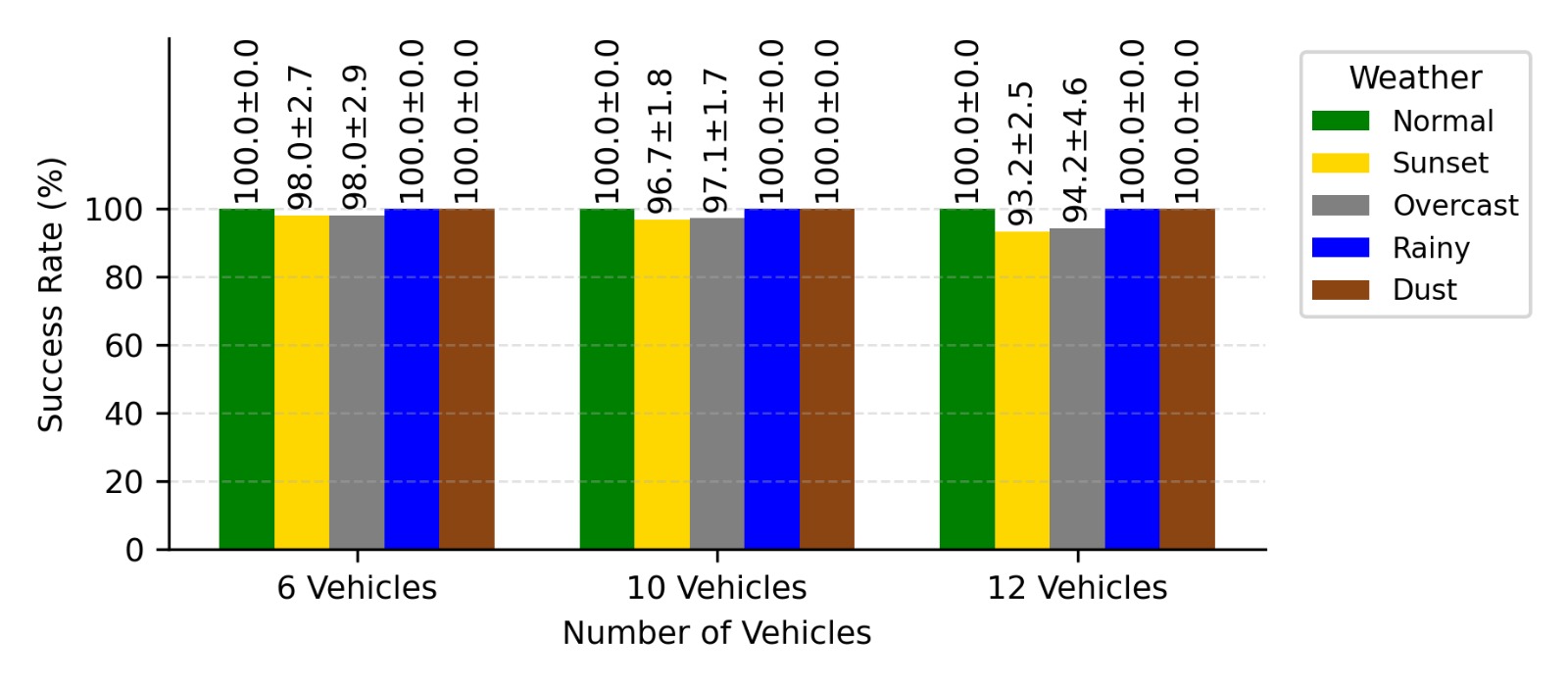}
    \caption{Success rate under different weather conditions
    }
    \label{fig:weather_plot}
\end{figure}

To further assess the robustness of the proposed end-to-end framework, we evaluate its performance under diverse weather conditions simulated in CARLA, as illustrated in Fig.~\ref{fig:testbed}. The corresponding results are presented in Fig.~\ref{fig:weather_plot}. The framework maintains consistently high performance across all tested conditions, demonstrating strong generalization capability and achieving near-perfect success rates even in visually challenging scenarios.

%\begin{figure}
%    \centering
    %\includegraphics[width=0.4\linewidth]{Weather_conditions.png}
    %\caption{Illustration of the various weather conditions simulation in the CARLA simulator.}
    %\label{fig:weather_conditions}
%\end{figure}

\section{Conclusion} \label{sec:conclusion}

%The experimental results directly validate the core design considerations of the proposed testbed. 
We present a mixed-reality, hardware-in-the-loop testbed integrating the CARLA simulator with physical robots offering scalability. The testbed leverages ROS as a middleware and incorporates vehicle-to-vehicle (V2V) connectivity, enabling mult-agent research using CAVs. The perception, planning, and control results demonstrate that the testbed enables end-to-end evaluation of AD algorithms under realistic sensing and environmental conditions. Moreover, the built-in data acquisition pipeline supports online learning and fine-tuning of AV systems, while the multi-agent configuration with synchronized virtual and physical vehicles provides a realistic framework for CAV research.   %Collectively, these make the testbed a powerful platform for advancing both AV and CAV research.
% CARLA provides photorealistic rendering and diverse environmental conditions, supporting realistic perception-driven evaluation of AV systems. 
%Our presented safety-guaranteed algorithm integrating perception, planning, and CBF-based control, offers a safety-certified platform for implementing, testing and evaluation of AV/CAV algorithms. %Specifically, our algorithm is based on online self-supervised learning which fine-tunes the controller, optimizing the joint performance of all CAVs in the network. 
Experimental results validate the efficacy of the proposed algorithm and the capabilities of the HIL platform, demonstrating its ability to bridge the gap between simulation and real-world deployment. Finally, our presented end-to-end control framework with CBFs provides a platform for safely testing and evaluating AV/CAV algorithms using the testbed. In the future, we aim to make our testbed public and remotely accessible for research on AVs, CAVs and mixed traffic involving HDVs. A major open challenge in this setting is accurately capturing the complex and often uncertain interactions between HDVs and CAVs, while still guaranteeing that all safety specifications are satisfied.
%{\color{blue} Mention that a major challenge is capturing interactions between HDVs and CAVs so as to still guarantee that all safety specifications are guaranteed. Cite one or two of our own recent papers but also those of others working on mixed traffic.}

%TODO: Sensor fusion details and related works

\bibliographystyle{IEEEtran}
\bibliography{example}  % .bib

@InProceedings{pmlr-v70-amos17a,
  title     = {{O}pt{N}et: Differentiable Optimization as a Layer in Neural Networks},
  author    = {Brandon Amos and J. Zico Kolter},
  booktitle = {Proceedings of the 34th International Conference on Machine Learning},
  series    = {Proceedings of Machine Learning Research},
  volume    = {70},
  pages     = {136--145},
  year      = {2017},
  month     = {06--11 Aug},
  publisher = {PMLR}
}

@misc{chen_multi-view_2017,
	title = {Multi-View 3D Object Detection Network for Autonomous Driving},
	author = {Chen, Xiaozhi and Ma, Huimin and Wan, Ji and Li, Bo and Xia, Tian},
	date = {2017-06-22},
	eprinttype = {arxiv},
	eprint = {1611.07759},
}

@inproceedings{dosovitskiy2017carla,
  title={CARLA: An open urban driving simulator},
  author={Dosovitskiy, Alexey and Ros, German and Codevilla, Felipe and Lopez, Antonio and Koltun, Vladlen},
  booktitle={Proceedings of the 1st Annual Conference on Robot Learning},
  pages={1--16},
  year={2017},
  organization={PMLR}
}

@misc{AgilexLimoPro,
  author       = {Agilex Robotics},
  title        = {{LIMO PRO: ROS\,/\,Gazebo Mobile Robot Platform with NVIDIA Orin Nano, T-mini Pro LiDAR, Orbbec Dabai Depth Camera}},
  howpublished = {\url{https://global.agilex.ai/products/limo-pro}},
  note         = {Accessed: Sep. 11, 2025},
  year         = {2025}
}

@misc{MathWorksRoadRunner,
  author       = {MathWorks, Inc.},
  title        = {{RoadRunner: Design 3D Scenes for Automated Driving Simulation}},
  howpublished = {\url{https://www.mathworks.com/products/roadrunner.html}},
  note         = {Accessed: Sep. 11, 2025},
  year         = {2025}
}

@article{paull2017duckietown,
  title={Duckietown: an open, inexpensive and flexible platform for autonomy education and research},
  author={Paull, Liam and Tani, Jacopo and Ahn, Heejin and Alonso-Mora, Javier and Carlone, Luca and Cap, Michal and De Wagter, Christophe and Dosovitskiy, Alexey and Dub{\'e}, Renaud and Espinoza, Victor and others},
  journal={2017 IEEE International Conference on Robotics and Automation (ICRA)},
  pages={1497--1504},
  year={2017},
  organization={IEEE}
}

@article{pickem2017robotarium,
  title={The Robotarium: A remotely accessible swarm robotics research testbed},
  author={Pickem, Daniel and Glotfelter, Paul and Wang, Li and Mote, Mark and Ames, Aaron D and Feron, Eric and Egerstedt, Magnus},
  journal={2017 IEEE International Conference on Robotics and Automation (ICRA)},
  pages={1699--1706},
  year={2017},
  organization={IEEE}
}

@inproceedings{scheffe2018scaled,
  title={A scaled experiment platform to study interactions between humans and CAVs},
  author={Scheffe, Patrick and Alrifaee, Bassam},
  booktitle={2018 IEEE Intelligent Vehicles Symposium (IV)},
  pages={1939--1944},
  year={2018},
  organization={IEEE}
}

@article{stager2018scaled,
  title={A scaled smart city for experimental validation of connected and automated vehicles},
  author={Stager, Adam and Bhan, Lavanya and Malikopoulos, Andreas and Zhao, Liuhui},
  journal={IFAC-PapersOnLine},
  volume={51},
  number={9},
  pages={130--135},
  year={2018},
  publisher={Elsevier}
}

@inproceedings{vargas2019development,
  title={Development of sensors testbed for autonomous vehicles},
  author={Vargas, Jorge M and Alsweiss, Suleiman and Jernigan, Michael and Amin, Abu Bony and Brinkmann, Marius and Santos, Joshua and Razdan, Rahul},
  booktitle={2019 IEEE 9th Annual Computing and Communication Workshop and Conference (CCWC)},
  pages={0574--0579},
  year={2019},
  organization={IEEE}
}

@article{abouchakra2025realissim,
  title={Real-is-Sim: Bridging the Sim-to-Real Gap with a Dynamic Digital Twin for Real-World Robot Policy Evaluation},
  author={Abou-Chakra, Jad and Sun, Lingfeng and Rana, Krishan and May, Brandon and Schmeckpeper, Karl and Minniti, Maria Vittoria and Herlant, Laura},
  journal={arXiv preprint arXiv:2504.03597},
  year={2025}
}

@inproceedings{chen2019selective,
  title={Selective sensor fusion for neural visual-inertial odometry},
  author={Chen, Changhao and Rosa, Stefano and Miao, Yishu and Lu, Chris Xiaoxuan and Wu, Wei and Markham, Andrew and Trigoni, Niki},
  booktitle={Proceedings of the IEEE/CVF Conference on Computer Vision and Pattern Recognition},
  pages={10542--10551},
  year={2019}
}

@INPROCEEDINGS{Ames_01,
  author={Ames, Aaron D. and Galloway, Kevin and Grizzle, J. W.},
  booktitle={2012 IEEE 51st IEEE Conference on Decision and Control (CDC)},
  title={Control lyapunov functions and hybrid zero dynamics}, 
  year={2012},
  volume={},
  number={},
  pages={6837-6842},
  doi={10.1109/CDC.2012.6426229}}

@article{mokhtarian2024survey,
  title={A Survey on Small-Scale Testbeds for Connected and Automated Vehicles and Robot Swarms},
  author={Mokhtarian, Armin and Xu, Jianye and Scheffe, Patrick and Kloock, Maximilian and Sch{\"a}fer, Simon and Bang, Heeseung and Le, Viet-Anh and Ulhas, Sangeet and Betz, Johannes and Wilson, Sean and Berman, Spring and Paull, Liam and Prorok, Amanda and Alrifaee, Bassam},
  journal={arXiv preprint arXiv:2408.14199},
  year={2024}
}

@inproceedings{ahmad2025hierarchical,
  title     = {HMARL-CBF: Hierarchical Multi-Agent Reinforcement Learning with Control Barrier Functions for Safety-Critical Autonomous Systems},
  author    = {Ahmad, H. M. Sabbir and Sabouni, Ehsan and Wasilkoff, Alexander and Budhraja, Param and Guo, Zijian and Zhang, Songyuan and Fan, Chuchu and Cassandras, Christos G. and Li, Wenchao},
  booktitle = {Advances in Neural Information Processing Systems},
  year      = {2025},
}

@inproceedings{hyldmar2019fleet,
  title        = {A Fleet of Miniature Cars for Experiments in Cooperative Driving},
  author       = {Hyldmar, Nicholas and He, Yijun and Prorok, Amanda},
  booktitle    = {2019 IEEE International Conference on Robotics and Automation (ICRA)},
  year         = {2019},
  pages        = {3238--3244},
  organization = {IEEE}
}

@article{chalaki2021research,
  title={A Research and Educational Robotic Testbed for Real-time Control of Emerging Mobility Systems: From Theory to Scaled Experiments},
  author={Chalaki, Behdad and Beaver, Logan E. and Mahbub, A. M. Ishtiaque and Bang, Heeseung and Malikopoulos, Andreas A.},
  journal={arXiv preprint arXiv:2109.05370},
  year={2021}
}

@book{xiao2023safe,
  title={Safe autonomy with control barrier functions: theory and applications},
  author={Xiao, Wei and Cassandras, Christos G and Belta, Calin},
  year={2023},
  publisher={Springer}
}

@article{treiber2000congested,
  title={Congested traffic states in empirical observations and microscopic simulations},
  author={Treiber, Martin and Hennecke, Ansgar and Helbing, Dirk},
  journal={Physical review E},
  volume={62},
  number={2},
  pages={1805},
  year={2000},
  publisher={APS}
}

\end{document}